\documentclass[10pt,twocolumn,letterpaper]{article}

\usepackage{cvpr}
\usepackage{times}
\usepackage{epsfig}
\usepackage{graphicx}
\usepackage{amsmath}
\usepackage{amssymb}
\usepackage{subfigure}
\usepackage{multirow}
\usepackage[linesnumbered,vlined,ruled]{algorithm2e}

\SetCommentSty{mycommfont}

\SetKwInput{KwInput}{Input}                
\SetKwInput{KwOutput}{Output}              

\newcommand{\ourcluster}{CoDIAL\xspace}
\newcommand{\oursampling}{Uniform Entropy\xspace} 

\usepackage[pagebackref=true,breaklinks=true,letterpaper=true,colorlinks,bookmarks=false]{hyperref}

\cvprfinalcopy 

\def\cvprPaperID{9468} 

\ifcvprfinal\fi
\begin{document}

\title{Low-Budget Label Query through Domain Alignment Enforcement}

\author{Jurandy Almeida$^1$, Cristiano Saltori$^2$, Paolo Rota$^2$, and Nicu Sebe$^2$\\[1ex]
$^1$Instituto de Ci\^{e}ncia e Tecnologia, Universidade Federal de S\~{a}o Paulo -- UNIFESP\\
12247-014, S\~{a}o Jos\'{e} dos Campos, SP -- Brazil\\
{\tt\small jurandy.almeida@unifesp.br}\\[1ex]
$^2$Dept. of Information Engineering and Computer Science, University of Trento -- UniTn\\
38123, Trento, TN -- Italy\\
{\tt\small \{cristiano.saltori, paolo.rota, niculae.sebe\}@unitn.it}\\
}

\maketitle

\begin{abstract}
Deep learning revolution happened thanks to the availability of a massive amount of labelled data which have contributed to the development of models with extraordinary inference capabilities. Despite the public availability of a large quantity of datasets, to address specific requirements it is often necessary to generate a new set of labelled data.  Quite often, the production of labels is costly and sometimes it requires specific know-how to be fulfilled. In this work, we tackle a new problem named \textit{low-budget label query} that consists in suggesting to the user a small (low budget) set of samples to be labelled, from a completely unlabelled dataset, with the final goal of maximizing the classification accuracy on that dataset. In this work we first improve an Unsupervised Domain Adaptation (UDA) method to better align source and target domains using consistency constraints, reaching the state of the art on a few UDA tasks. Finally, using the previously trained model as reference, we propose a simple yet effective selection method based on uniform sampling of the prediction consistency distribution, which is deterministic and steadily outperforms other baselines as well as competing models on a large variety of publicly available datasets.
\end{abstract}


\section{Introduction}
\label{sec:intro}

The large amount of data generated on daily basis led to the blooming of performing data hungry models capable of impressive results. However, there are many niche applications where such production is not ideal and producing a desirable amount of data to satiate such models is costly and unfeasible. In biomedical imaging, sharing data is critical due to privacy issues; industrial applications such as quality control or predictive maintenance often require a specific type of data which is often rare, unbalanced and necessitates professional expertise to be correctly labelled. For these reasons, having an adequate amount of labelled data to solve a specific problem can be still costly and not always affordable. Labelling is therefore crucial, not only in the quantity of data produced but also for the quality of the labelled samples.

In this work, we tackle the problem of having a very limited budget for data labelling by defining a methodology that deterministically identifies a convenient small subset of data to be manually labelled thus becoming usable for practical applications. To the best of our knowledge this is an entirely novel problem that is relevant for a wide range of tasks from Active Learning \cite{beluch2018power,lewis1994sequential} where an initial pool of labelled data is required to perform the iterative active query or more applicative use cases where a reasonable amount of data is available but budget for labelling is limited. Deciding which samples have to be prioritized for labelling in a new unseen dataset is a non-trivial task. In this work, we aim at finding a method that is able to select a pool of samples from an unlabelled dataset to be manually labelled in order to maximize the classification performance in such dataset. We will refer to this problem as \textit{low-budget label query}. 

The proposed approach is divided in three main phases namely Domain Adaptation, Budget Sampling, and Classification. In the first phase, we adapt the source trained model through Unsupervised Domain Adaptation (UDA)~\cite{chen2019progressive,cicek2019unsupervised,ganin2015unsupervised,long2018conditional,rozantsev2018beyond} in order to align the feature distributions of the source and target datasets. This domain agnostic model is then used as reference to compute the classification uncertainty of each sample in the target dataset. This phase is quite critical, we need a model capable of producing a good ranking list of samples according to the reliability of the prediction. Such reliability, however, is not a warranty of good decision considering how neural networks tend to predict with over-confidence even when they suggest a wrong prediction~\cite{zou2019confidence}. To this end we took inspiration from AutoDIAL \cite{carlucci2017autodial}, where the distributions of source and target are aligned towards the separate computation of the batch normalization layers' statistics and the model is trained using two simple losses, one supervised and one unsupervised for the source and the target, respectively. In this work we modified AutoDIAL replacing the unsupervised entropy loss with a cost function that maximizes the consistency between pristine target images and their randomly perturbed versions. This modification improves the accuracy of the UDA on most datasets we considered. We will refer to this method as \ourcluster (Consistency DIAL).

In the second phase, we first sample a fixed budget of examples from the most to the last confidently classified examples using the adapted model, then we ask the user to label the selected examples obtaining the low-budget labelled pool.
The adapted model is therefore used to indicate a balanced distribution of representative samples aiming at maximizing the accuracy in a low budget labelling scenario. 
It is well known that high confidence samples are likely to be located close to the center of the marginal distribution of a given class and therefore they are well representing the class itself, however low confidence samples are situated in a grey area between classes and are therefore possibly defining the shape of the marginal distribution of the class. Our intuition is that a balanced contribution of the two types of sampling shall be more representative than either one of them. For this reason we propose to sample uniformly the examples on the confidence metrics distribution.
In the final phase, the source model is fine-tuned using the selected sampled as human annotated thus obtaining a model that is able to perform well on the target dataset.
In this paper we show that this solution outperforms the other strategies in this scenario.

The contributions of our paper can be summarized as follows: 
\begin{itemize}
    \item An improved method for UDA, named \ourcluster, which improves the method proposed in \cite{carlucci2017autodial} and has comparable results with the state of the art on many publicly available datasets;
    \item As far as we know, this is the first work investigating the \textit{low-budget label query} problem enabling methods such as active learning or self-paced learning to have something better than a \textit{random selection} as the initial pool of labelled samples;
    \item A method based on uniform sampling of the confidence metrics distribution, which is deterministic and steadily outperforms other baselines.  
\end{itemize}

\section{Related Works}
\label{sec:related}

Although our work faces a novel yet interesting problem that often affects the computer vision and, more in general, the machine learning communities, the proposed method relates to other problems that have been treated before in literature. Here we provide an overview of such problems highlighting differences and affinities with our solution.

\textbf{Unsupervised Domain Adaptation (UDA).}
The main goal of \textit{Unsupervised Domain Adaptation}~\cite{chen2019progressive,cicek2019unsupervised,ganin2015unsupervised,long2018conditional,rozantsev2018beyond} is to reduce the discrepancy or shift between the distributions of a source and a target domain without target labels. A first class of methods focus on aligning source and target distributions through \textit{domain confusion} either by introducing a feature alignment loss \cite{tzeng2014deep} or by exploiting adversarial learning \cite{tzeng2017adversarial}. Following this direction, Volpi~\etal~\cite{volpi2018adversarial} improve the adversarial alignment by introducing a generative source feature sampler strategy aiming at augmenting source data in the feature space. Differently, other methods achieve domain alignment by introducing domain dependent batch normalization layers \cite{carlucci2017autodial} or domain dependent batch whitening layers \cite{roy2019unsupervised}.
More recently, Sun~\etal~\cite{sun2019unsupervised} introduce the concept of \textit{self-supervision} in UDA aligning source and target features in a shared domain-invariant feature space.
In our work, we address the problem of low-budget label query, proposing an UDA method to align the distributions of the source and target domains, obtaining a domain-invariant confidence metric that allows for a better sampling for the labeling procedure.

\textbf{Self-Paced Learning (SPL).} 
Bengio~\etal~\cite{bengio2009curriculum} introduced the concept of curriculum learning in which a model is trained starting from the easiest samples and then is gradually introduced with more complex samples along the training. The concept has been improved by Kumar~\etal~\cite{kumar2010self} introducing the paradigm of \textit{Self-Paced Learning} (SPL). The SPL model includes a weighted loss term on all samples and a more general regularization imposed on sample weights. Weights are optimized during training and the pace parameter is updated, allowing the model to discover new samples in a self-paced way. This training methodology has been applied on different topics such as video event detection~\cite{jiang2015self} and object detection~\cite{sangineto2018self} to name a few. Although our work is not directly related to SPL, the sampling strategy facilitates labelling the most representative samples, providing a balanced representation from the simplest to the most difficult samples.

\begin{figure*}[t!]
    \centering
    \includegraphics[width=\linewidth]{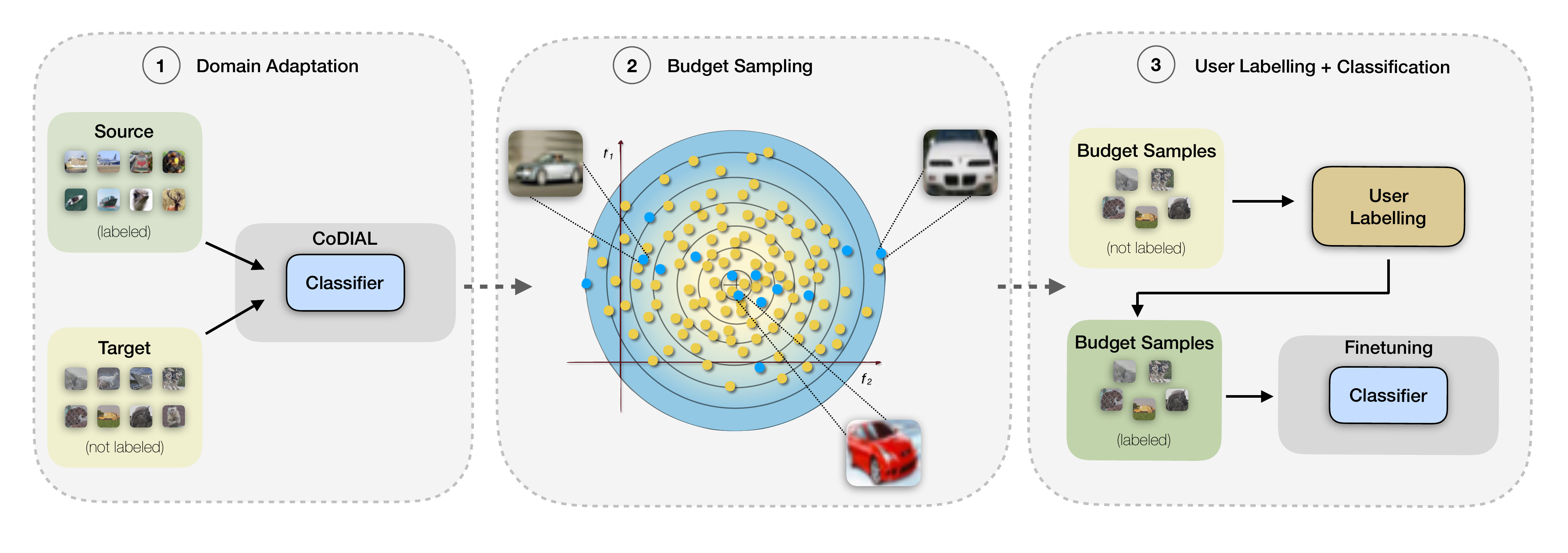}
    \caption{Overall view of the proposed method.}
    \label{fig:scheme}
\end{figure*}

\textbf{Active Learning (AL).}
The concept of \textit{Active Learning} refers to an iterative learning technique that foresees the active involvement of the user to provide information about the most uncertain samples. While the user is already directly involved in the labelling of the highest entropy samples in the seminal work of Lewis~\etal~\cite{lewis1994sequential}, the first explicit definition of AL is presented in the technical report of Settles \cite{settles2009active}. More recently, Beluch~\etal~\cite{beluch2018power} propose a more reliable uncertainty measure computed on an ensemble of networks instead of a single one.
The issue of the \textit{cold start} is a well-known problem in the initialization step of AL methods \cite{gao2019consistency,houlsby2014cold,konyushkova2017learning} that is often partially solved by randomly sampling the initial set of labeled data.
While the goal of our work is not to compete with AL, the sampling strategy proposed in our pipeline can be used to further alleviate the cold start problem providing a better initialization strategy compared to the random sampling. 

\textbf{Low-budget labelling.}
While deep learning methods often require a large amount of labelled data to show reasonable performance, data labeling still remains an expensive activity that often requires highly specialized labor to be performed. In spite of its relevance, the low-budget label query task has been considered only by a few works. Recently, in \cite{gao2019consistency} the minimization of the labeling cost is partially tackled by applying \textit{Semi-Supervised Learning} in an AL approach to obtain better candidates for the user labelling procedure. Although the task of labeling cost minimization is related to low-budget label query, our pipeline differs from \cite{gao2019consistency} in two major and fundamental points. First, our method aims at finding the most representative fixed number of samples to be labeled among all unlabeled data while in \cite{gao2019consistency}, despite of the improved candidate proposal, the entire dataset is iteratively labeled. This is a fundamental difference since our method requires the user to label only a fixed number of data reducing drastically the labor required by such activity. Finally, in \cite{gao2019consistency} the initial budget of samples is randomly initialized bringing back the cold start problem.

\section{Methodology}
\label{sec:method}

In this section, we describe our approach to perform low-budget label query. In particular, in Section~\ref{sec:da}, we describe the proposed UDA method while in Section~\ref{sec:budget}, we futher describe our proposed budget sampling strategy. For clarity, an overview of our approach is provided in Fig.~\ref{fig:scheme}.

\subsection{Consistency-based Domain Alignment (\ourcluster)}
\label{sec:da}
As long as no information is provided on the target dataset, we assume to have a set of labelled data $\mathcal{S} = \left \{ (x_1^s, y_1^s), \dots, (x_n^s, y_n^s) \right \}$ sharing the same set of labels with our unlabelled dataset $\mathcal{T} = \left \{ x_1^t, \dots, x_m^t \right \}$. In the first phase, we perform UDA between the labelled dataset $\mathcal{S}$ and the unlabelled dataset $\mathcal{T}$. This procedure is meant to train a model to perform well on a source dataset but at the same time aligning the distributions with the target dataset. To carry out this task, we took inspiration from AutoDIAL~\cite{carlucci2017autodial}, a method that exploits separate batch normalization statistics for source and target while training the model weights jointly. The loss proposed in~\cite{carlucci2017autodial} is composed of two components, one supervised for $\mathcal{S}$ and one unsupervised for $\mathcal{T}$.  
In our model, we use the same supervised loss $\mathcal{L}_s(\theta)$, which is the sparse cross-entropy in Eq.~\eqref{eq:supervised}:
\begin{equation}\label{eq:supervised}
    \mathcal{L}_s(\theta) = -\frac{1}{n}\sum_{i=1}^{n}\log f_{s}^{ \theta }\left ( y_{i}^{s}; x_{i}^{s} \right )
\end{equation}
\noindent where $f_{s}^{ \theta }\left ( y_{i}^{s}; x_{i}^{s} \right )$ is the probability of $x_i^s$ to be assigned to class $y_i^s$ and $n$ is the number of source samples.

The unsupervised loss proposed in AutoDIAL minimizes the entropy of the target samples in order to force the model to decide more confidently. The related term of the loss refers to the entropy of the data distribution in $\mathcal{T}$ as shown in Eq.~\eqref{eq:entropyloss}:
\begin{equation}\label{eq:entropyloss}
    \mathcal{L}_e(\theta) = -\frac{1}{m}\sum_{i=1}^{m}\sum_{y\in \mathcal{Y}}f_{t}^{ \theta }\left ( y; x_{i}^{t} \right ) \log f_{t}^{ \theta }\left ( y; x_{i}^{t} \right )
\end{equation}
where $m$ is the number of samples in the target batch and $\mathcal{Y}$ is the set of target labels.

\begin{figure*}[h!]
    \centering
    \subfigure[]{\includegraphics[width=0.3\textwidth]{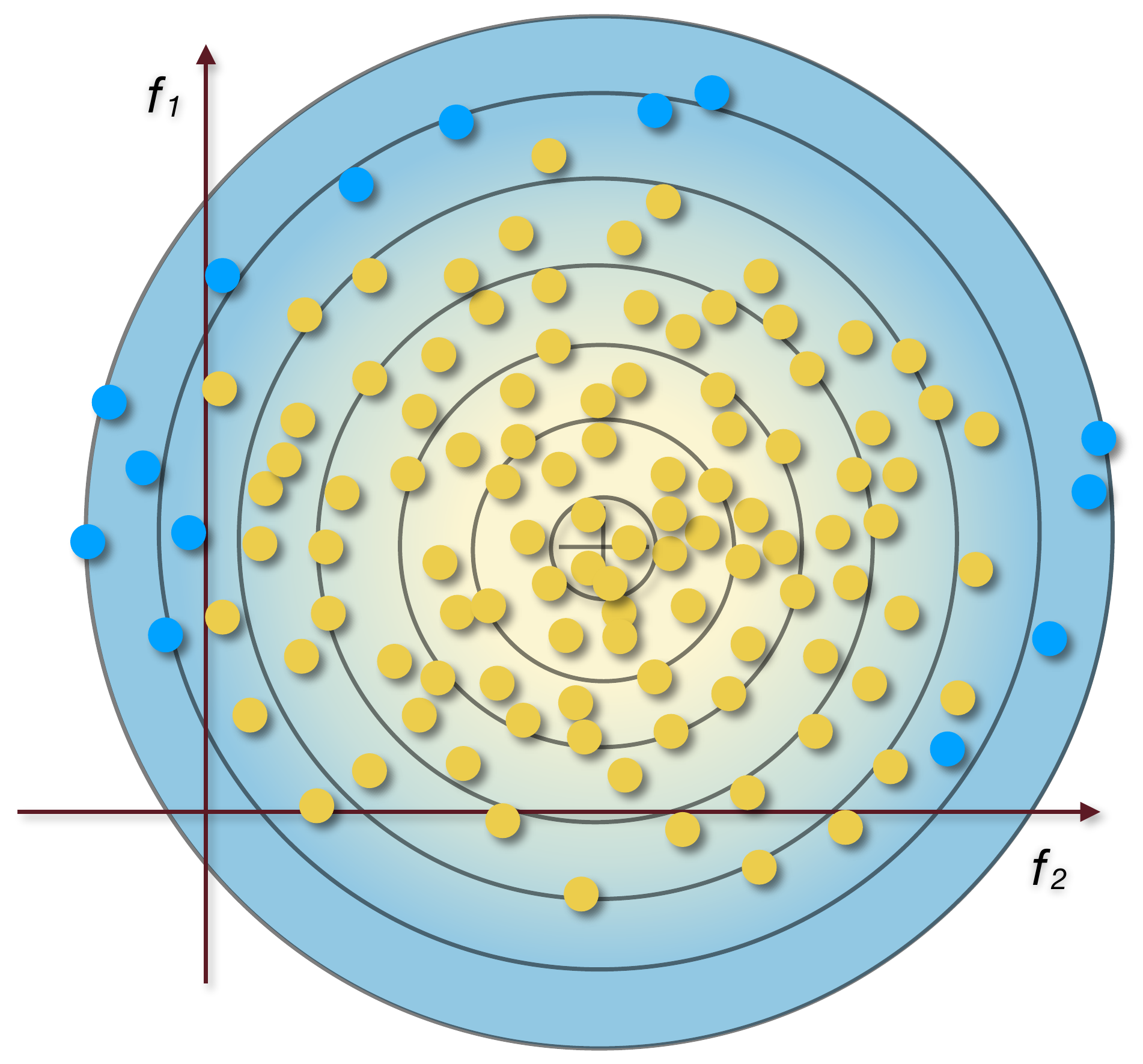}} 
    \subfigure[]{\includegraphics[width=0.3\textwidth]{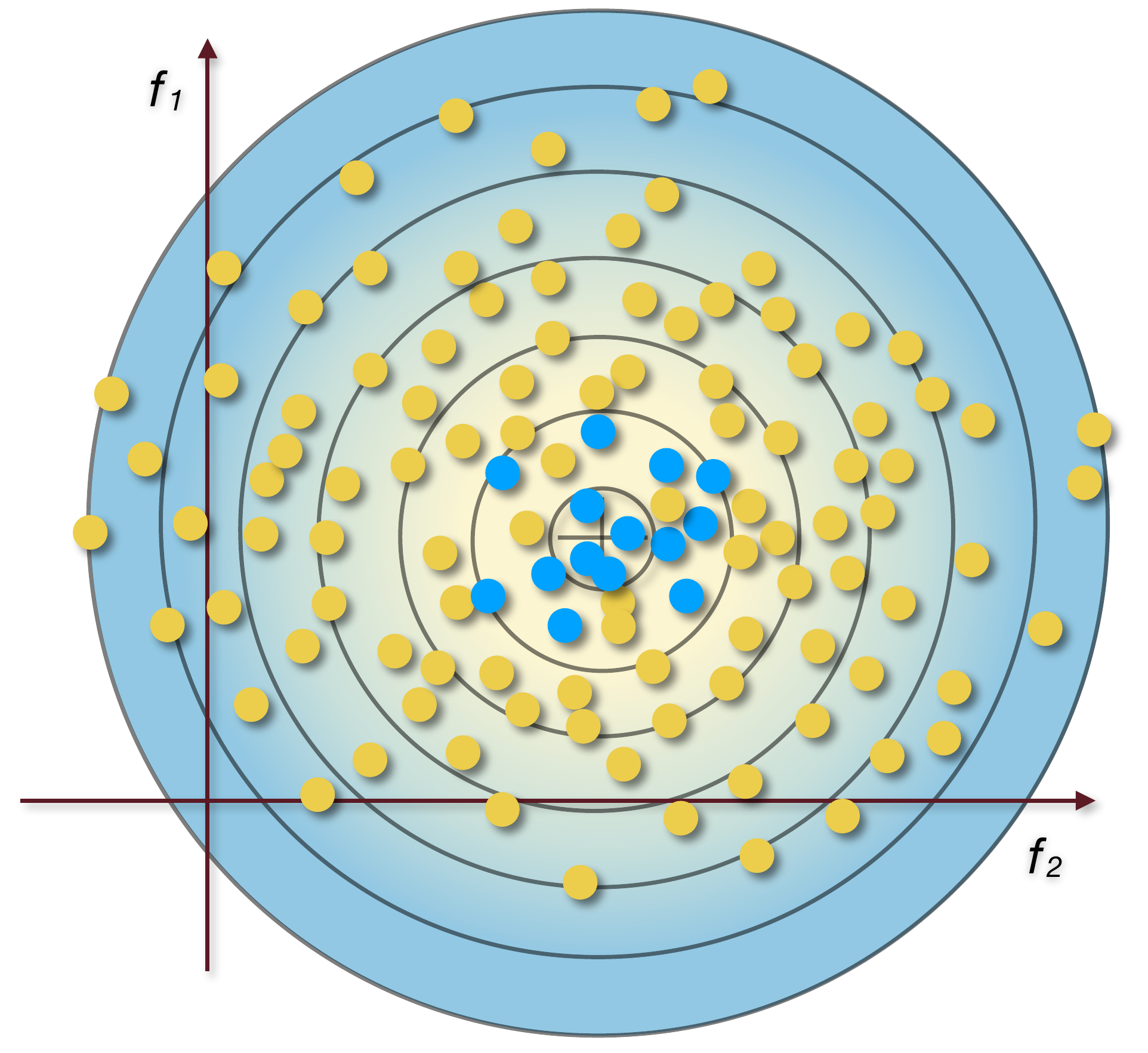}} 
    \subfigure[]{\includegraphics[width=0.3\textwidth]{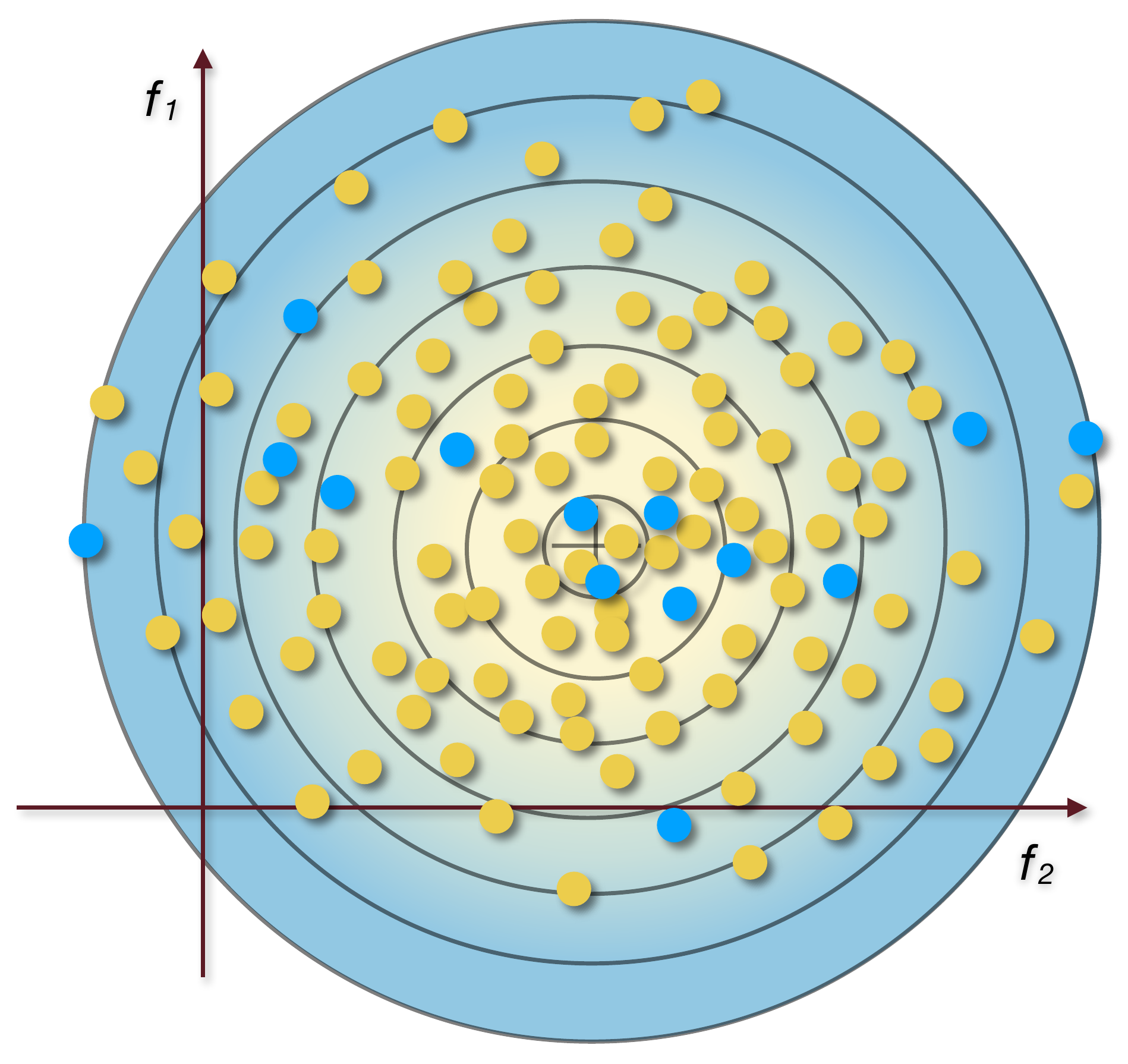}} 
    \caption{(a) Toprank, (b) Minrank and (c) Uniform sampling strategies. The latter allows to include samples of increasing degree of confidence and difficulty.}
    \label{fig:sampling}
\end{figure*}

In order to improve the robustness of our model, we introduce a consistency constraint in the loss function. Consistency loss is common in unsupervised learning and it is often enforced by minimizing the distance between a pristine and a perturbed version of the same image~\cite{gao2019consistency,sajjadi2016regularization}. In our formulation, we opted for a KL-divergence to compute the feature distance between the two images as shown in Eq.~\eqref{eq:consistencyloss}:
\begin{equation}\label{eq:consistencyloss}
    \mathcal{L}_c(\theta) = \frac{1}{m}\sum_{i=1}^{m} \sum_{y\in \mathcal{Y}} f_{t}^{ \theta }\left ( y;x_{i}^{t} \right ) \log \frac{f_{t}^{ \theta }\left ( y;x_{i}^{t} \right )}{f_{t}^{ \theta }\left ( y;\tilde{x}_{i}^{t} \right )}
\end{equation}

\noindent where $\tilde{x}_{i}^{t}$ is the perturbed version of the original image ${x}_{i}^{t}$.

After unifying Eq.~\eqref{eq:entropyloss} and Eq.~\eqref{eq:consistencyloss}, the overall unsupervised loss $\mathcal{L}_u(\theta)$ in our model can be reduced to the cross-entropy\footnote{Please refer to the Supplementary Materials for a detailed demonstration.} between $x_{i}^t$ and $\tilde{x}_{i}^t$ as shown in Eq.~\eqref{eq:unsupervised}.

\begin{equation}\label{eq:unsupervised}
    \mathcal{L}_u(\theta) = -\frac{1}{m}\sum_{i=1}^{m} \sum_{y\in \mathcal{Y}} f_{t}^{ \theta }\left ( y;x_{i}^{t} \right ) \log f_{t}^{ \theta }\left ( y;\tilde{x}_{i}^{t} \right )
\end{equation}

The final loss $\mathcal{L}(\theta)$ of our adaptation model is a weighted sum of $\mathcal{L}_s(\theta)$ and $\mathcal{L}_u(\theta)$:

\begin{equation}\label{eq:overallloss}
    \mathcal{L}(\theta) = \mathcal{L}_s(\theta) + \lambda \mathcal{L}_u(\theta)
\end{equation}

\noindent where $\lambda$ is an hyper parameter weighting the contribution of the unsupervised term.

\subsection{Budget Samples Selection (BSS)}
\label{sec:budget}
The selection of the budget samples is the central part of this work. Such samples, once manually labelled, should lead to a convenient subset of data optimizing the classification accuracy. 
To this end, we propose a simple yet effective approach that leverages the previously adapted model (see Section \ref{sec:da}) to produce a reliable method that computes inference confidence. To measure the classification confidence of a sample, a large variety of approaches is available in literature such as consistency \cite{gao2019consistency}, ensemble consistency \cite{beluch2018power} and euclidean distance between k-centers \cite{sener2017active} to name a few. 
In our work, we took in exam entropy as being a simple but yet effective and widely used metrics~\cite{gao2019consistency,carlucci2017autodial,roy2019unsupervised} for classification confidence.
Although the decision of the confidence metrics has an important role in the current problem, the sampling policy assumes the same importance. 
Our proposal takes the advantages of two different sampling policies: Toprank and Minrank. These methods consist in selecting those samples that have an higher and lower value of entropy therefore including the most and the least uncertain samples respectively.
In Toprank it is asked to the user to label the hardest samples, those on the edge of the marginal distribution where intuitively most complex samples are located. However a model trained only with edge samples lacks of the characterization of the most representative samples that are selected using the Minrank method. Given these observations, we propose to merge these approaches sampling the examples using an uniform policy w.r.t. the confidence metrics. 

\begin{figure}[h!]
    \centering
\includegraphics[width=0.48\textwidth]{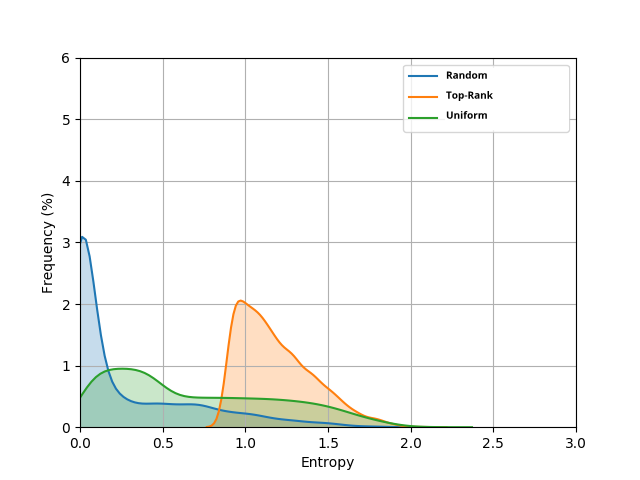}
    \caption{The distribution of entropy of the selected samples of CIFAR10 for the model trained on STL10 with CoDIAL. The example shows three sampling strategies, (blue) the random sampling (resembling consistency metrics distribution), (orange) the top rank entropy strategy and (green) the proposed approach based on uniform sampling.}
    \label{fig:distros}
\vspace{-2mm}
\end{figure}

A visual representation of the three aforementioned sampling strategies are shown in Figure~\ref{fig:sampling}, while in Figure \ref{fig:distros} are shown three examples of confidence distribution where the sampling is done using Random, Toprank and Uniform policy. For sake of clarity, Minrank has been omitted because it would follow a near-dirac distribution close to 0.

For clarity, we report the overall budget selection strategy in Algorithm~\ref{alg_cross_obj_det}.

\begin{algorithm}[h]
\SetKwInOut{Input}{Input}
\SetKwInOut{Output}{Output}
\small{
\caption{Our budget selection algorithm\label{alg_cross_obj_det}}
\KwData{$\mathcal{T}$ -- the target training set of unlabeled samples}
\Input{
\begin{itemize}
\item $S$ -- a difficulty scoring function, e.g., entropy given by $S(x_{i}^{t}) = -\sum_{y\in \mathcal{Y}}f^{ \theta }\left ( y; x_{i}^{t} \right ) \log f^{ \theta }\left ( y; x_{i}^{t} \right )$, where $f^{ \theta }$ is a model used to compute the probability of $x^t$ to be assigned to the pseudo-label $y$
\item $k$ -- the number of samples to be selected for the budget
\end{itemize}}
\Output{$\mathcal{B}$ -- the set of samples selected for the budget}
\For{$i \gets 1$ \KwTo $m$}
{
    $s_{i} \gets S(x_{i}^{t})$\;
}
$s_{min} \gets \mathrm{min}(\{s_{1}, \dots, s_{m}\})$\;
$s_{max} \gets \mathrm{max}(\{s_{1}, \dots, s_{m}\})$\;
\For{$i \gets 1$ \KwTo $m$}
{
    $b_{i} \gets \left \lfloor k \times \frac{s_{i} - s_{min}}{s_{max} - s_{min}} \right \rfloor + 1$\;
}
$\mathcal{B} \gets \emptyset$\;
\While{$|\mathcal{B}| < k$}
{
    $\mathcal{Q} \gets \emptyset$ \tcp*{$Q$ is a priority queue}
    \For{$j \gets 1$ \KwTo $k$}
    {
        $max_{j} \gets 0$; $s_{max_{j}} \gets -\infty$\;
        \For{$i \gets 1$ \KwTo $m$}
        {
            \If{$b_{i} = j$ \textrm{\bf and} $s_{i} > s_{max_{j}}$ \textrm{\bf and} $x_{i}^{t} \notin \mathcal{B}$}
            {
                $max_{j} \gets i$; $s_{max_{j}} \gets s_{i}$\;
            }
        }
        \If{$max_{j} > 0$}
        {
            \textsc{Insert}$(\mathcal{Q}, \langle s_{max_{j}}, max_{j} \rangle)$\;
        }
    }
    \While{$|\mathcal{B}| < k$ \textrm{\bf and} $|\mathcal{Q}| > 0$}
    {
        $\langle s, i \rangle \gets $ \textsc{Extract-Max}$(\mathcal{Q})$\;
        $\mathcal{B} \gets \mathcal{B} \bigcup \{ x_{i}^{t} \}$\;
    }
}
}
\end{algorithm}

\section{Experiments}
\label{sec:exps}
In this section, we provide details about the experimental setup adopted in order to evaluate our approach as well as we report the obtained results.
A rigorous and extensive experimental evaluation was performed for each of the tasks foreseen by our approach (see  Section~\ref{sec:method}): (1) UDA and (2) BSS.
The experiments are conducted on both small and large-scale datasets and the proposed method is compared with state-of-the-art approaches. 
In Section \ref{sec:ablation}, discussion of the achievements is included to further analyze the relation between the confidence and the consistency of a model and its impact on the budget samples selection.

\subsection{Datasets}
\label{sec:datasets}
In order to assess the effectiveness of our approach, we conducted experiments on seven publicly-available datasets widely used in UDA tasks:

\textbf{CIFAR-10$\leftrightarrow$STL.}
The CIFAR-10 dataset~\cite{krizhevsky2009learning} is composed of 60000 RGB images, with a size of $32\times32$. CIFAR-10 comes already balanced with 6000 images for each class and it is divided into a training set of 50000 images and a test set of 10000 images. Inspired to the latter, the STL dataset~\cite{coates2011analysis} consists of a total of 5000 images where each of the 10 classes is represented by 500 $96\times96$ RGB images. In our experiments, we remove the non-overlapping classes \textit{frog} from CIFAR-10 and \textit{monkey} from STL.

\textbf{MNIST$\leftrightarrow$USPS.}
The MNIST dataset~\cite{lecun1998gradient} is composed of 60000 training and 10000 test gray-scale images of handwritten digits in a range from 0 to 9. Each image has a fixed-size of $28\times28$ pixels.
Similar to MNIST, the USPS dataset~\cite{hull1994database} is a smaller handwritten digits dataset composed of 7291 training and 2007 testing gray-scale $16\times16$ images. In our experiments, we exploit the already zero-padded $32\times32$ MNIST and USPS images from~\cite{paszke2017automatic}.

\textbf{SVHN$\rightarrow$MNIST.}
The Street View House Number (SVHN) dataset~\cite{netzer2011reading} is a real-world MNIST-like digits dataset composed of 73257 training and 26032 testing RGB $32\times32$ images. As for MNIST, also in SVHN classes are in the range from 0 to 9.
Despite of similarities, the unbalanced number of per-class images, the often severe changes of illumination and the non-centered digits depicting represent a significant domain shift.

\textbf{Office-31.}
The Office-31~\cite{saenko2010office31} dataset is a standard benchmark in the domain adaptation literature. This dataset is composed of 4652 images collected from Amazon.com or taken from an office environment using a Webcam or a DSLR camera and with varying lighting and pose changes. Those images comprise 31 classes from three different domains: Amazon (\textbf{A}), DSRL (\textbf{D}) and Webcam (\textbf{W}). 

\begin{table*}
    \centering
    \begin{tabular}{c|cc|c|c|c|c}
        \hline
        \hline
                         & Source & CIFAR-10 & STL      & SVHN  & MNIST & USPS \\
        \textit{Methods} & Target & STL      & CIFAR-10 & MNIST & USPS  & MNIST \\
        \hline
        Source only      &        &  60.35   &  51.88   & 60.10 & 78.90 & 57.10\\ 
        \hline
        AutoDIAL~\cite{carlucci2017autodial} &  & 79.10 & 70.15 & 89.12 & 97.96 & 97.51 \\
        DWT~\cite{roy2019unsupervised}       &  & 79.75 & 71.18 & 97.75 & \textbf{99.09} & 98.79 \\
        \textbf{CoDIAL (Ours)} & & \textbf{81.06} & \textbf{71.48} & \textbf{98.32} & 97.51 & \textbf{98.88} \\ 
        \hline
        Target only      &        &  67.75   &  88.86   & 99.50 & 96.50 & 99.20 \\ 
        \hline
        \hline
    \end{tabular}
    \caption{Accuracy (\%) on the CIFAR-10$\leftrightarrow$STL, MNIST$\leftrightarrow$USPS, and SVHN$\rightarrow$MNIST datasets and comparison with state-of-the-art methods.}
    \label{tab:uda:smalldatasets}
\end{table*}

\begin{table*}[!htb]
    \centering
    \begin{tabular}{c|ccccccc|c}
        \hline
        \hline
                         & Source &   A   &   A   &   D   &   D   &   W   &   W   &        \\
        \textit{Methods} & Target &   D   &   W   &   A   &   W   &   A   &   D   &  Avg.  \\
        \hline
        ResNet-50~\cite{he2016resnet} & & 
        81.7 & 76.5 & 65.2 & 97.7 & 65.6 & \textbf{100.0} & 81.1 \\
        AutoDIAL~\cite{carlucci2017autodial} & & 
        87.2 & 88.1 & 65.9 & 99.0 & 63.8 & \textbf{100.0} & 84.0 \\
        \textbf{CoDIAL (Ours)} & & 
        \textbf{89.4} & \textbf{90.4} & \textbf{68.0} & \textbf{99.1} & \textbf{66.2} & \textbf{100.0} & \textbf{85.5} \\
        \hline
        \hline
    \end{tabular}
    \caption{Accuracy (\%) on the Office-31 dataset with Resnet-50 as base network and comparison with state-of-the-art methods.}
    \label{tab:uda:office31}
\end{table*}

\textbf{Office-Home.}
The Office-Home~\cite{venkateswara2017officehome} dataset is a large-scale benchmark widely-used for testing domain adaptation methods. This dataset is composed of 15500 images collected from several search engines and online image directories. Those images are distributed among 65 object categories and are divided into 4 distinct domains: Art (\textbf{Ar}), Clipart (\textbf{Cl}), Product (\textbf{Pr}) and Real World (\textbf{Rw}). 

Unlike the CIFAR-10$\leftrightarrow$STL, MNIST$\leftrightarrow$USPS, and SVHN$\rightarrow$MNIST datasets, there are no predefined train and test splits in the Office-31 and Office-Home datasets.
For that reason, all possible combinations of source and target domains are evaluated using a full protocol setting~\cite{gong2013fullprotocol}, where the entire source (with labels) and target (without labels) data are used for training and all the target samples are used for testing.

\subsection{Implementation Details}
\label{sec:experimental_setup}
For a fair comparison, we adopt the network described in~\cite{french2017self} for the CIFAR-10$\leftrightarrow$STL experiments and the architecture in~\cite{ganin2016domain} for the digits experiments (\ie, MNIST$\leftrightarrow$USPS and SVHN$\rightarrow$MNIST).
In order to apply \ourcluster, we follow~\cite{carlucci2017autodial} and add source and target batch normalization layers with the difference that we apply an additional batch normalization layer to separately normalize perturbed target batches.
For the UDA tasks, the network is trained from scratch for 120 epochs with a mini-batch size of $32$ by using the Adam optimization algorithm~\cite{kingma2014adam} with a weight decay of $5\times10^{-4}$ and an initial learning rate of $1\times10^{-3}$ with scheduled decay of $0.1$ at the epochs 50 and 90.
For the BSS tasks, the network is initialized with the weights obtained in the corresponding UDA tasks and then fine-tuned for 50 epochs using the Adam optimizer with an initial learning rate of $1\times10^{-4}$, step-decay with a decay rate of $0.1$ every 10 epochs, mini-batch size of $32$ and weight decay regularization of $5\times10^{-4}$.

For the Office-31 and Office-Home experiments, we follow~\cite{long2018conditional} and use a ResNet-50~\cite{he2016resnet} architecture.
First, all the batch normalization layers of ResNet-50 are replaced with \ourcluster layers, which uses three batch normalization layers to separately normalize source, pristine target, and perturbed target images.
Then, we initialize the network with ImageNet pretrained weights replacing the output layer by a fully-connected layer with $C$ output logits and randomly initialized weights, where $C$ is the number of classes (\ie, $C = 31$ for Office-31 and $C = 65$ for Office-Home).
For the UDA tasks, we train the network from scratch for 60 epochs with a mini-batch size of $20$ by using the SGD~(Stochastic Gradient Descent) optimization algorithm with a weight decay of $5\times10^{-4}$, an initial learning rate of $1\times10^{-2}$ for the randomly initialized parameters of the output layer and $1\times10^{-3}$ for the rest of the trainable parameters of the network. 
Step-decay is used to reduce the initial learning rates by a factor of 10 at the epoch 54.

During the training of \ourcluster, we apply transformations to pristine target batches in order to obtain perturbed target batches.
For that, we use the same perturbations adopted in~\cite{roy2019unsupervised}: for the CIFAR-10$\leftrightarrow$STL, Office-31, and Office-Home experiments, we apply random affine transformations (as proposed in~\cite{french2017self}), Gaussian blur ($\sigma = 0.1$), random horizontal flips with a probability of 50\%, and random crop with padding of 4 pixels, whereas for the digits experiments we adopt the same perturbations, with exception of the horizontal flips.
The $\lambda$ parameter of Eq.~\ref{eq:overallloss} is tuned for each experiment. For the visualization of the learning curves and accuracy values along training we have exploited the W\&B platform~\cite{wandb}.

\subsection{Unsupervised Domain Adaptation}
\label{sec:UDA_results}
In Table~\ref{tab:uda:smalldatasets} is reported a comparison among results obtained by our \ourcluster with those recently reported in~\cite{roy2019unsupervised} for the CIFAR-10$\leftrightarrow$STL, MNIST$\leftrightarrow$USPS, and SVHN$\rightarrow$MNIST datasets.
For a fair comparison, we considered only the results in~\cite{roy2019unsupervised} obtained without the application of data augmentation on the source dataset.
Notice that \ourcluster succeed in most of the UDA tasks, outperforming all the baselines and reaching state-of-the-art performance on four out of five proposed benchmark.

\begin{table*}[!htb]
    \centering
    \begin{tabular}{c|ccccccccccccc|c}
        \hline
        \hline
                         & Source &   Ar   &   Ar   &   Ar   &   Cl   &   Cl   &   Cl   &   Pr   &   Pr   &   Pr   &   Rw   &   Rw   &   Rw   &        \\
        \textit{Methods} & Target &   Cl   &   Pr   &   Rw   &   Ar   &   Pr   &   Rw   &   Ar   &   Cl   &   Rw   &   Ar   &   Cl   &   Pr   &  Avg.  \\
        \hline
        ResNet-50~\cite{he2016resnet} & & 
        34.9 & 50.0 & 58.0 & 37.4 & 41.9 & 46.2 & 38.5 & 31.2 & 60.4 & 53.9 & 41.2 & 59.9 & 46.1 \\
        DAN~\cite{roy2019unsupervised} & & 
        43.6 & 57.0 & 67.9 & 45.8 & 56.5 & 60.4 & 44.0 & 43.6 & 67.7 & 63.1 & 51.5 & 74.3 & 56.3 \\
        DANN~\cite{roy2019unsupervised} & & 
        45.6 & 59.3 & 70.1 & 47.0 & 58.5 & 60.9 & 46.1 & 43.7 & 68.5 & 63.2 & 51.8 & 76.8 & 57.6 \\
        JAN~\cite{roy2019unsupervised} & & 
        45.9 & 61.2 & 68.9 & 50.4 & 59.7 & 61.0 & 45.8 & 43.4 & 70.3 & 63.9 & 52.4 & 76.8 & 58.3 \\
        CDAN-RM~\cite{long2018conditional} & & 
        49.2 & 64.8 & 72.9 & 53.8 & 63.9 & 62.9 & 49.8 & 48.8 & 71.5 & 65.8 & 56.4 & 79.2 & 61.6 \\
        CDAN-M~\cite{long2018conditional} & & 
        50.6 & 65.9 & 73.4 & 55.7 & 62.7 & 64.2 & 51.8 & 49.1 & 74.5 & 68.2 & 56.9 & 80.7 & 62.8 \\
        DWT~\cite{roy2019unsupervised} & & 
        \textbf{50.8} & 72.0 & 75.8 & 58.9 & 65.6 & 60.2 & 57.2 & \textbf{49.5} & 78.3 & 70.1 & \textbf{55.3} & 78.2 & 64.3 \\
        \textbf{CoDIAL (Ours)} & & 
        45.3 & \textbf{73.2} & \textbf{77.5} & \textbf{59.8} & \textbf{70.2} & \textbf{71.7} & \textbf{58.0} & 41.4 & \textbf{78.7} & \textbf{70.6} & 49.0 & \textbf{81.3} & \textbf{64.7} \\
        \hline
        \hline
    \end{tabular}
    \caption{Accuracy (\%) on the Office-Home dataset with Resnet-50 as base network and comparison with state-of-the-art methods.}
    \label{tab:uda:officehome}
\vspace{-5mm}
\end{table*}

\begin{figure*}
    \centering
    \subfigure[]{\includegraphics[width=0.48\textwidth]{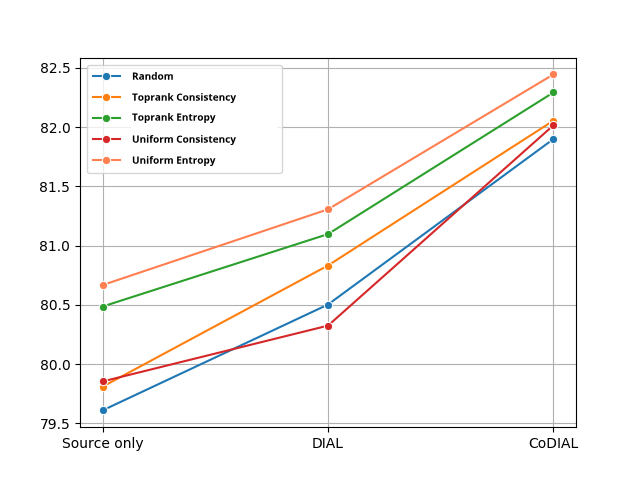}} 
    \subfigure[]{\includegraphics[width=0.48\textwidth]{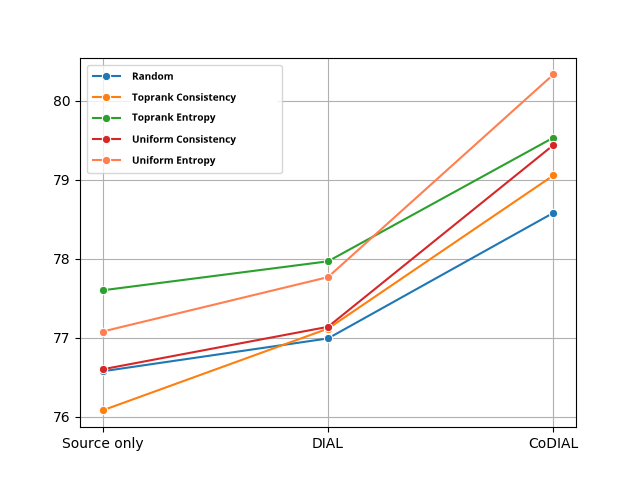}} 

    \caption{Performances of all the sampling strategies with $k=10\%$ of $|\mathcal{T}|$ with Source only model, AutoDIAL and \ourcluster in the task CIFAR-10$\rightarrow$STL (a), STL$\rightarrow$CIFAR-10 (b). For more graphs refer to Supplementary Materials.     }
    \label{fig:tasks:cifar-stl}
\vspace{-2mm}
\end{figure*}

Table~\ref{tab:uda:office31} compares the results achieved by \ourcluster on the Office-31 dataset. As a reference, we also report the results obtained by the ResNet-50~\cite{he2016resnet} model pretrained on the ImageNet~\cite{russakovsky2015imagenet} dataset and finetuned on the source data.
As we can observe, \ourcluster outperformed AutoDIAL in all the UDA tasks of the Office-31 dataset, yielding a gain in the average performance of 1.5\%.

Finally, Table~\ref{tab:uda:officehome} compares the results obtained by \ourcluster with those recently reported in~\cite{roy2019dwt} for the Office-Home dataset.
Observe that our \ourcluster performed better than all the compared methods in all the UDA tasks of the Office-Home dataset, except for those where Clipart (\textbf{Cl}) is the target domain.
Indeed, \ourcluster is more effective than the previous approaches, achieving the best average performance and improving accuracy by up to 6.7\% in the UDA tasks where Clipart (\textbf{Cl}) is the source domain and with a considerable margin on a great part of the other benchmarks.

\subsection{Budget Samples Selection}
\label{sec:BSS_results}

In order to investigate the benefits of our solution, we compare it with a variety of widely-used sampling strategies, namely:

\noindent\textbf{Random.} This method simply selects a budget randomly sampling from the target training data.

\noindent\textbf{Toprank Entropy.} This strategy selects the examples with the highest value of entropy as budget samples.

\noindent\textbf{Toprank Consistency.} In this strategy the consistency score presented in~\cite{gao2019consistency} is used as the difficulty scoring function $S$ to select the budget samples. Similarly to Toprank Entropy, the samples with highest values of consistency are selected.

\noindent\textbf{Uniform Consistency.} In this strategy, the consistency score of~\cite{gao2019consistency} instead of the entropy is used as the difficulty scoring function $S$ in the algorithm of our approach (see Algorithm~\ref{alg_cross_obj_det}).

\begin{table*}
\centering
\resizebox{0.99\textwidth}{!}{
\begin{tabular}{c|c|cc|c|c|c|c}
\hline
\hline
\multirow{2}{*}{\textit{Model}} & \multirow{2}{*}{\textit{Sampler}} & Source & CIFAR-10 & STL & SVHN & MNIST & USPS \\
& & Target & STL & CIFAR-10 & MNIST & USPS & MNIST \\
\hline
\multirow{5}{*}{\rotatebox{90}{\parbox{1cm}{\centering Source only}}} 
& Random & & 79.61 $\pm$ 0.52 & 76.58 $\pm$ 0.29 & 98.65 $\pm$ 0.07 & 93.43 $\pm$ 0.45 & 98.84 $\pm$ 0.08 \\
& Toprank Entropy & & 80.49 & 77.60 & 98.51 & 89.24 & 98.45 \\
& Toprank Consistency~\cite{gao2019consistency} & & 79.81 $\pm$ 0.32 & 77.08 $\pm$ 0.37 & 98.66 $\pm$ 0.09 & 93.21 $\pm$ 0.36 & 98.30 $\pm$ 0.19 \\
& Uniform Consistency & & 79.85 $\pm$ 0.38 & 76.08 $\pm$ 0.24 & 98.74 $\pm$ 0.06 & 93.49 $\pm$ 0.33 & 98.87 $\pm$ 0.05 \\
& \oursampling  & & 80.67 & 76.60 & 98.58 & 93.37 & 98.78 \\
\hline
\multirow{5}{*}{\rotatebox{90}{AutoDIAL~\cite{carlucci2017autodial}}} 
& Random & & 80.50 $\pm$ 0.44 & 76.99 $\pm$ 0.18 & 98.98 $\pm$ 0.06 & 96.28 $\pm$ 0.17 & 98.94 $\pm$ 0.06 \\
& Toprank Entropy & & 81.10 & 77.97 & 99.46 & 96.31 & 99.42 \\
& Toprank Consistency~\cite{gao2019consistency} & & 80.83 $\pm$ 0.32 & 77.12 $\pm$ 0.35 & 99.16 $\pm$ 0.07 & 96.29 $\pm$ 0.11 & 99.04 $\pm$ 0.06 \\
& Uniform Consistency & & 80.33 $\pm$ 0.23 & 77.14 $\pm$ 0.42 & 99.18 $\pm$ 0.05 & 96.25 $\pm$ 0.17 & 99.17 $\pm$ 0.06 \\
& \oursampling  & & 81.31 & 77.77 & 99.38 & 96.41 & 99.43 \\
\hline
\multirow{5}{*}{\rotatebox{90}{\parbox{1cm}{\centering \ourcluster (Ours)}}}
& Random & & 81.90 $\pm$ 0.36 & 78.58 $\pm$ 0.33 & 99.10 $\pm$ 0.07 & 97.58 $\pm$ 0.16 & 99.28 $\pm$ 0.06 \\
& Toprank Entropy & & 82.29 & 79.53 & \textbf{99.52} & \textbf{97.91} & 99.40 \\
& Toprank Consistency~\cite{gao2019consistency} & & 82.05 $\pm$ 0.28 & 79.06 $\pm$ 0.23 & 99.49 $\pm$ 0.03 & 97.78 $\pm$ 0.08 & 99.42 $\pm$ 0.02 \\
& Uniform Consistency & & 82.02 $\pm$ 0.30 & 79.44 $\pm$ 0.30 & 99.46 $\pm$ 0.03 & 97.59 $\pm$ 0.14 & \textbf{99.44 $\pm$ 0.03} \\
& \textbf{\oursampling (ours)} & & \textbf{82.44} & \textbf{80.33} & 99.48 & \textbf{97.91} & 99.43 \\
\hline
\hline
\end{tabular}
}
\caption{Accuracy (\%) on the target test set using a budget with $k = 10\%$ of $|\mathcal{T}|$ for the CIFAR-10$\leftrightarrow$STL, MNIST$\leftrightarrow$USPS, and SVHN$\rightarrow$MNIST datasets.}
\label{tab:smalldatasets:results10}
\end{table*}

\begin{table*}
\centering
\resizebox{0.99\textwidth}{!}{
\begin{tabular}{c|c|cc|c|c|c|c}
\hline
\hline
\multirow{2}{*}{ \textit{Model} } & \multirow{2}{*}{ \textit{Sampler} } & Source & CIFAR-10 & STL & SVHN & MNIST & USPS  \\
& & Target & STL & CIFAR-10 & MNIST & USPS & MNIST \\
\hline
\multirow{5}{*}{\rotatebox{90}{\parbox{1cm}{\centering Source only}}} 
& Random & & 77.89 $\pm$ 0.36 & 68.13 $\pm$ 0.38 & 94.71 $\pm$ 0.35 & 84.17 $\pm$ 1.02 & 95.78 $\pm$ 0.17 \\
& Toprank Entropy & & 77.57 & 67.59 & 91.68 & 69.27 & 83.78\\
& Toprank Consistency~\cite{gao2019consistency} & & 77.43 $\pm$ 0.24 & 67.36 $\pm$ 0.44 & 95.14 $\pm$ 0.58 & 82.93 $\pm$ 3.60 & 93.26 $\pm$ 0.76 \\
& Uniform Consistency & & 77.57 $\pm$ 0.43 & 67.81 $\pm$ 0.46 & 95.24 $\pm$ 0.28 & 77.67 $\pm$ 1.06 & 95.85 $\pm$ 0.24 \\
& \oursampling  & & 78.46 & 67.63 & 95.47 & 82.7 & 95.09\\
\hline
\multirow{5}{*}{\rotatebox{90}{AutoDIAL~\cite{carlucci2017autodial}}} 
& Random & & 79.59 $\pm$ 0.29 & 71.36 $\pm$ 0.52 & 98.44 $\pm$ 0.07 & 95.58 $\pm$ 0.15 & 98.41 $\pm$ 0.08 \\
& Toprank Entropy & & 79.74 & 71.94 & 94.79 & 95.82  & 98.80\\
& Toprank Consistency~\cite{gao2019consistency} & & 79.40 $\pm$ 0.13 & 71.45 $\pm$ 0.46 & 98.51 $\pm$ 0.09 & 95.79 $\pm$ 0.09 & 98.42 $\pm$ 0.06 \\
& Uniform Consistency & & 79.57 $\pm$ 0.13 & 71.52 $\pm$ 0.47 & 98.53 $\pm$ 0.11 & 95.57 $\pm$ 0.15 & 98.51 $\pm$ 0.09 \\
& \oursampling  & & 79.76 & 71.91 & 95.83 & 95.77 &98.78\\
\hline
\multirow{5}{*}{\rotatebox{90}{\parbox{1cm}{\centering \ourcluster (Ours)}}} 
& Random & & 80.77 $\pm$ 0.25 & 72.86 $\pm$ 0.42 & 98.39 $\pm$ 0.11 & 97.50 $\pm$ 0.13 & 98.95 $\pm$ 0.04 \\
& Toprank Entropy & & 80.63 & 72.88 & 98.88 & 97.51 & \textbf{99.17}\\
& Toprank Consistency~\cite{gao2019consistency} & & 80.47 $\pm$ 0.20 & 72.73 $\pm$ 0.41 & 98.68 $\pm$ 0.04 & 97.62 $\pm$ 0.10 & 99.06 $\pm$ 0.04 \\
& Uniform Consistency & & 80.63 $\pm$ 0.22 & \textbf{73.44 $\pm$ 0.34} & 98.85 $\pm$ 0.07 & 97.52 $\pm$ 0.10 & 99.12 $\pm$ 0.08 \\
& \textbf{\oursampling (ours)} & & \textbf{81.14} & 73.24 & \textbf{98.95} & \textbf{97.66}& 99.11\\
\hline
\hline
\end{tabular}
}
\caption{Accuracy (\%) on the target test set using a budget with $k = 1\%$ of $|\mathcal{T}|$ for the CIFAR-10$\leftrightarrow$STL, MNIST$\leftrightarrow$USPS, and SVHN$\rightarrow$MNIST datasets.}
\label{tab:smalldatasets:results1}
\end{table*}

In the following experiments, we evaluate the accuracy obtained on the target test set after fine-tuning the models trained in the UDA tasks reported in Section~\ref{sec:UDA_results} on a target sampled subset obtained by using the aforementioned methods.
In order to have statistically sound results, we repeat 10 times the experiments for the non-deterministic sampling strategies (\ie, Random, Toprank Consistency, and Uniform Consistency) and report the mean and the standard deviation.
To provide a full view of the jointly advantages of our approach, we evaluate the performance of the sampling strategies together with AutoDIAL~\cite{carlucci2017autodial} instead of our \ourcluster and also without using domain alignment (referred as \textit{Source only} in tables).
Due to space limitations, we chose to report the results only for the CIFAR-10$\leftrightarrow$STL, MNIST$\leftrightarrow$USPS, and SVHN$\rightarrow$MNIST datasets while the results for the Office-31 and Office-Home datasets are reported in the supplementary materials.

Table~\ref{tab:smalldatasets:results10} presents the results obtained with a budget size of $k=10\%$ of $|\mathcal{T}|$. 
In Table~\ref{tab:smalldatasets:results1}, additional results obtained with $k=1\%$ of $|\mathcal{T}|$ are presented. 
For sake of clarity, an overall view of the performances of all the sampling methods with $k=10\%$ of $|\mathcal{T}|$ among with Source only model, AutoDIAL and \ourcluster is presented in Figure~\ref{fig:tasks:cifar-stl}.
Notice that \oursampling performs better than the other sampling strategies in most of the UDA tasks, yielding  a gain in the accuracy of up to 1.75\%.

By comparing the results from Tables~\ref{tab:smalldatasets:results10} and~\ref{tab:smalldatasets:results1}, it is possible to note that even with a very limited budget sample size of $k=1\%$ of $|\mathcal{T}|$, the fine-tuned classifier improves the previous performance obtained both with (\textit{AutoDIAL} and \textit{\ourcluster}) or without domain alignment (\textit{Source only}).

Despite \oursampling performs reasonably well even with Source only or AutoDIAL~\cite{carlucci2017autodial}, the best results are obtained by jointly exploiting both \ourcluster and \oursampling.
Indeed, our \ourcluster is essential for the performance of our \oursampling, as discussed in the next section. 

\begin{figure*}[!htb]
    \centering
    \includegraphics[width=\textwidth]{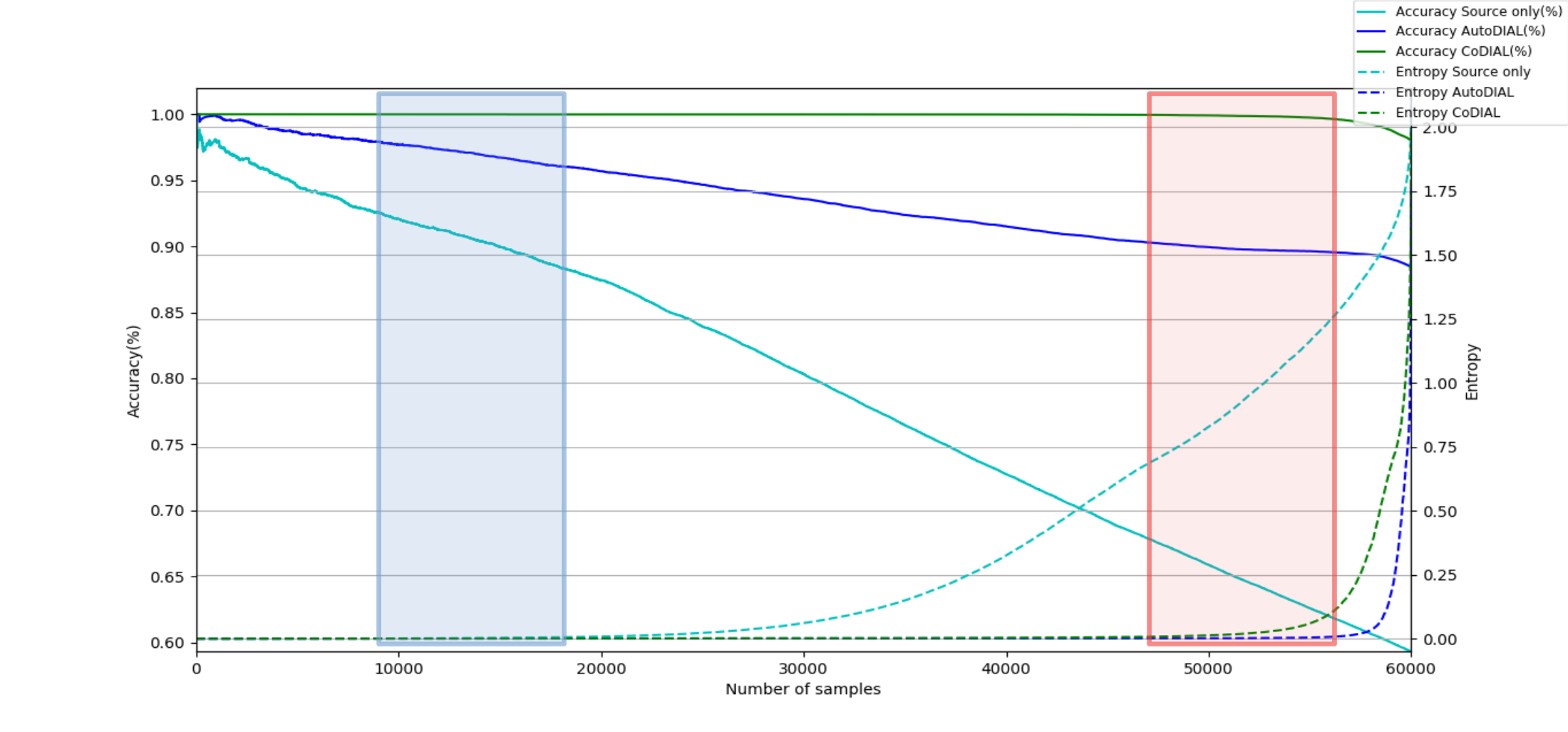}
    \caption{The entropy-accuracy relation for three different models evaluated on the SVHN$\rightarrow$MNIST dataset.}
    \label{fig:entrop-acc}
\end{figure*}

\subsection{Discussion}
\label{sec:ablation}
In this section, we give an intuition of the importance of the \ourcluster in combination with our sampling strategy discussing the relation between confidence and consistency of a model and its impact on the BSS task. In specific, we focus our investigation on the high entropy regime, which is more interesting due to affecting samples close to the edge of the marginal distribution and therefore near the decision border.

Confidence is related to the certainty of a model about its predictions whereas consistency is related to its capacity to encourage predictions to be similar under small perturbations of inputs or network parameters~\cite{athiwaratkun2019consistency}.
We argue that a possible ideal budget sample has low confidence but high consistency, meaning it is hardly well classified no matter what is the transformation applied to the sample.
To support our assumption, we analyze the relation between the confidence of a sample prediction and its impact on the performance of a model.
To this end, entropy is used to assess the confidence of each sample prediction while accuracy is used to assess the model.
In Figure~\ref{fig:entrop-acc}, we show a plot relating the entropy of each sample and its impact on the accuracy.
In this plot, the x-axis refers to the number of budget samples, the y-axis on the right refers to the entropy of each sample and the y-axis on the left denotes the accuracy of the model.
The samples are sorted in an increasing order of their entropy values then each sample is selected according to its entropy value, i.e. the lowest is the first, the second lowest is the second, and so on (\ie, according to the x-axis, from left to right).
At each step, we take the next sample, including it into a subset with all the previously selected ones and compute the accuracy of the model for this subset (\ie, the lowest entropy samples).
In this way, the accuracy decreases if the selected sample is wrongly classified.

In Figure~\ref{fig:entrop-acc}, we compare the entropy-accuracy relation for the three models evaluated in Section~\ref{sec:BSS_results}: Source only, AutoDIAL~\cite{carlucci2017autodial}, and \ourcluster. 
Due to space limits, we choose to report and discuss only the results obtained for the SVHN$\rightarrow$MNIST dataset, however a similar behavior may be observed for all the other evaluated datasets.
In this figure, two important regions are highlighted: (1) one on the left (blue box) related to low entropy samples, and (2) one on the right (red box) concerning high entropy samples.
For all the three models, the entropy values (dashed lines) for the samples in the first region (blue box) are close to zero, therefore they all are very confident on their predictions.
Notice that, by increasing the number of samples, the accuracy of \ourcluster\ (solid line in green) remains unaffected, indicating that such samples were correctly classified.
However, the accuracy for Source only and AutoDIAL (solid lines in cyan and blue, respectively) is decreasing as more samples are considered, showing that they are over confident also when they take wrong decisions.
On the other hand, by analyzing the second region (red box), we can notice that AutoDIAL is more confident on its predictions than \ourcluster, since the entropy values start to raise for the samples predictions obtained with \ourcluster\ (dashed line in green) whereas they are still low for AutoDIAL (dashed line in blue).
In spite of that, a small drop can be observed in the accuracy of AutoDIAL (solid line in blue) while \ourcluster\ maintains a high accuracy (solid line in green) until the extremely highest entropy samples are reached. This further proves how CoDIAL is more robust w.r.t. AutoDIAL in classifying also high entropy, and often harder, samples.

The reasoning behind \oursampling is that a balanced contribution of both low and high confidence samples  benefits learning more than either only one of them.
A requirement for that is to have a model capable of producing reliable predictions.
In this sense, \ourcluster helps to hunt those samples where the model is overconfident but classify them wrongly, enabling \oursampling to better capture the marginal distribution of the classes.

\section{Conclusion}
\label{sec:conclusion}
Low-budget label query turned out to be a challenging problem that requires additional attention. Looking at the results obtained in this work we can assume that a uniform sampling of the confidence metrics tends to be convenient w.r.t. random sampling and other baselines. 

As an additional contribution, we have proposed an improved version of AuotDIAL that is reaching state-of-art performances on multiple benchmarks.
It is furthermore noticeable that our method can be easily plugged into a large variety of other tasks such as Self-Paced Learning or Active Learning applications. For Self-Paced or Incremental Learning, this work can be applied to improve the composition of exemplar sets~\cite{rebuffi2017icarl} while in Active Learning it can provide a more steady and reliable starting point helping to reduce the effect of the \textit{cold start problem}~\cite{konyushkova2017learning,houlsby2014cold}. The proposed method has the applicative limitation of requiring a source dataset that shares the same labels therefore as future work we would like to relax this constraint and consider other challenging applications such as open set domain adaptation and domain generalization. 

{\small
\bibliographystyle{ieee_fullname}
\bibliography{eccv2020}
}

\end{document}